\pdfoutput=1

\documentclass[11pt]{article}

\usepackage{emnlp2021}

\usepackage{times}
\usepackage{latexsym}
\usepackage{graphicx}
\usepackage{xcolor}
\usepackage{arabtex}
\usepackage{multirow}
\usepackage{makecell}
\usepackage{utf8}
\usepackage[utf8]{inputenc}
\setcode{utf8}
\usepackage[export]{adjustbox}
\usepackage{subcaption}

\usepackage[T1]{fontenc}

\usepackage[utf8]{inputenc}

\usepackage{microtype}

\usepackage{enumitem}

\newcommand{\squeezeup}{\vspace{-4mm}}
\newcommand{\squeezeupless}{\vspace{-2mm}}

\usepackage{color, colortbl}
\definecolor{LightCyan}{rgb}{0.95,1,1}
\definecolor{LightGrey}{rgb}{0.90,0.90,0.90}

\title{Automatic Expansion and Retargeting \\ of Arabic Offensive Language Training}

\author{Hamdy Mubarak, Ahmed Abdelali, Kareem Darwish and Younes Samih \\
  Qatar Computing Research Institute, Hamad Bin Khalifa University \\
  Doha, Qatar\\
  \texttt{\{hmubarak, aabdelali, kdarwish, ysamih\}@hbku.edu.qa} \\}

\begin{document}
\maketitle
\begin{abstract}
Rampant use of offensive language on social media led to recent efforts on automatic identification of such language.  Though offensive language has general characteristics, attacks on specific entities may exhibit distinct phenomena such as malicious alterations in the spelling of names.  In this paper, we present a method for identifying entity specific offensive language.  We employ two key insights, namely that replies on Twitter often imply opposition and some accounts are persistent in their offensiveness towards specific targets. Using our methodology, we are able to collect thousands of targeted offensive tweets. We show the efficacy of the approach on Arabic tweets with 13\% and 79\% relative F1-measure improvement in entity specific offensive language detection when using deep-learning based and support vector machine based classifiers respectively. Further, expanding the training set with automatically identified offensive tweets directed at multiple entities can improve F1-measure by 48\%.\newline

\end{list} 

\end{abstract}

\section{Introduction}
\label{introduction}

\begin{figure*}[!t]
    \centering
    \includegraphics[width=.75\columnwidth]{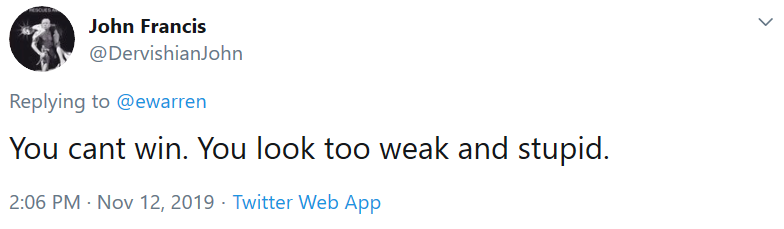}
    \includegraphics[width=.75\columnwidth]{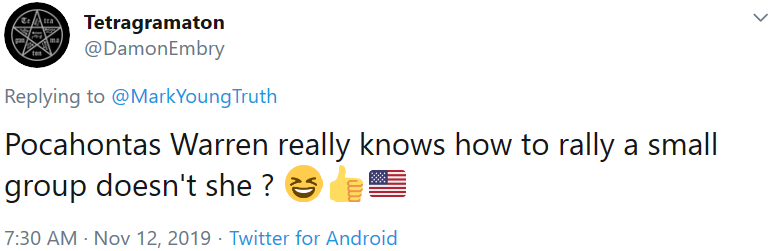}
    \caption{Example offensive tweets targeted towards Elizabeth Warren}
    \label{fig:sampleOffensiveTweetsElizabethWarren}
\end{figure*}

Social media has afforded users much freedom to express their thoughts with limited constraints on their language, which often includes name calling, offensive language, and hate speech.  Though such language may have general characteristics, targeted offensive language may exhibit distinct characteristics that may involve minor modifications in the spelling of the names of targets or the use of normally innocuous expressions in malicious ways.  For example, both tweets in Figure \ref{fig:sampleOffensiveTweetsElizabethWarren} are offensive towards the former Democratic presidential hopeful (2020 US elections) Elizabeth Warren.  Though the first is clearly offensive, the second uses a derogatory reference that is not explicitly offensive.  Hence, a generic classifier would likely detect the former and miss the latter. In this paper, we present a method for identifying entity specific offensive language. Doing so would allow us to retrain an offensive language classifier to detect such offensive language that is targeted towards a specific entity. We employ two key insights, namely that frequent replies to an entity on Twitter often imply opposition, and the appearance of offensive replies from a specific account towards an entity may indicate existence of other offensive replies from the same account.  Thus, given a target of interest, we identify accounts that frequently reply to that target.  Next, we run a generic offensive language classifier on the replies to identify the ``most offensive'' users based on either the number or percentage of their replies that are classified as offensive.  Then, we assume that the remaining replies of the most offensive users towards that target are also offensive.  The intuition behind this is based on the observation that users typically have persistent stances towards entities. Thus, negative stances lead to offensive replies.  
Then, we expanded the training set using the replies of offensive users.  This expansion is inspired by query expansion from information retrieval, where the top ranked documents in response to a query are assumed to be relevant and top terms from these documents are used to expand the query.  We show that the proposed expansion greatly enhances target specific classification effectiveness as well as offensive language detection in general.  \textcolor{black}{We focus in this paper on target specific offensive language in Arabic tweets.} 
The contributions of this paper are as follows:
\begin{itemize}
    \item We present a method for automatic expansion of offensive language training set.
    \item We show that the expanded training set leads to improved results when applied to targeted offensive language in specific and offensive language detection in general.
    \item \textcolor{black}{We show that our results hold when using two different classifiers, namely a Support Vector Machine (SVM) based classifier and a deep-leaning based classifier}.
\end{itemize}

\section{Related Work}
\label{related}
\label{sec:background}
Numerous
studies focused on 
the detection of offensive language, particularly hate speech in English \cite{agrawal2018deep,badjatiya2017deep,davidson2017automated,djuric2015hate,kwok2013locate,malmasi2017detecting,nobata2016abusive,yin2009detection}. \newcite{jay2008pragmatics} identified three categories of offensive speech, namely: \textbf{Vulgar}, which include explicit and rude sexual references, \textbf{Pornographic}, and \textbf{Hateful}, which includes offensive remarks concerning people's race, religion, country, etc.   
Most of these studies used supervised classification at either word level \cite{kwok2013locate}, character sequence level \cite{malmasi2017detecting,mubarak2019arabic}, and word embeddings \cite{djuric2015hate}. The studies used different classification techniques including using Na{\"i}ve Bayes \cite{kwok2013locate}, support vector machines \cite{malmasi2017detecting}, and deep learning \cite{agrawal2018deep,badjatiya2017deep,nobata2016abusive} classification. The accuracy of the aforementioned systems ranged between 76\% and 90\%.  

Work on Arabic offensive language detection is relatively new
.  \newcite{mubarak2019arabic} suggested that certain users are more likely to use offensive languages than others. 
\newcite{abozinadah2017detecting} used supervised classification based on a variety of features including user profile features, textual features, and network features, and reported an accuracy of nearly 90\%. \newcite{alakrot2018towards} used supervised classification based on word unigrams and n-grams to detect offensive language in YouTube comments, acheiving a precision of 88\%.  
\newcite{albadi2018they} focused on detecting religious hate speech using a recurrent neural network.  Since Arabic is a morphologically rich language and social media texts are often fraught with spelling mistakes, character n-gram text classification models generally yielded better results than word based models \cite{mubarak2019arabic}. 
Recently, shared tasks at SemEval 2020 \cite{zampieri2020semeval} and OSACT 2020 \cite{mubarak2020overview} included Arabic offensive language detection based on a corpus of 10,000 manually labeled tweets \cite{mubarak2020arabicOffArxiv}. In this paper, we utilize this corpus to train a baseline system. 


\section{Experiments} 
\label{classifiers}
\subsection{Experimental Setup}

As stated earlier, the purpose of this work is to improve target-specific Arabic offensive language classification on Twitter. We do so by automatically expanding our offensive language training set.  We employ a reasonably effective offensive language classifier to identify users who are antagonistic towards specific targets, and we use their tweets to expand our training set.  The main intuition is that users frequently use replies on Twitter to attack and discredit the target being replied to.  Thus, given Twitter users who reply frequently to a specific target and the classifier deems that a ``sufficient'' proportion or number of their replies are offensive, we assume that they are \textit{offensive users} and that the rest of their replies are offensive.  Hence, we are able to identify target specific expressions that are utilized in attacking the target that may not exist in our initial training set.  For example, 
one form of attacking the \textit{Aljazeera} news channel ( \<الجزيرة> --  ``Aljzyrp''\footnote{We use Buckwalter transliteration in this paper}) is to call it \textit{Alkhanzeera} (
\<الخنزيرة> --
``Alxnzyrp'' meaning a female pig/sow).  Both words rhyme as they share many letters.  Adding the tweets of offensive users to our training can help identify more target-specific offensive terms and may improve offensive language detection in general.\\
\textbf{Training Set:} 
For the training set, we used the Arabic offensive language shared task corpus that was used in SemEval 2020 \cite{zampieri2020semeval}, 
which is composed of 10,000 manually annotated tweets of which 1,915 are offensive.   \\
\begingroup
\renewcommand{\arraystretch}{1.3} 
\begin{table}[!t]
    \footnotesize
    \centering
    \begin{tabular}{c|r|r|r}
    Classifier & P & R & F1 \\ \hline
    SVM & \textbf{89.7} & 36.0 & 51.4 \\
    fastText & 83.9 &	\textbf{65.4} & \textbf{73.5} \\
    \end{tabular}
    \caption{Baseline results using cross validation}
    \label{tab:baselineCrossValResults}
\squeezeupless
\end{table}
\endgroup
\textbf{Baseline Experiments:} 
For baseline experiments, we performed 5-fold cross-validation on the training set using two classifiers. For the first, we used the SVMLight \cite{Joachims/99a} implementation of SVM classification with a linear kernel.  For the second, we used fastText \cite{bojanowski2016enriching}, which is a deep-learning based classifier, with a learning rate of 0.1 and 50 training epochs.  To overcome some of the challenges associated with Arabic text on social media, such as morphological complexity, use of dialects, and creative spellings, we used character n-grams ranging between 3 and 5 characters for both classifiers.  We also normalized all forms of \textit{alef} to \textit{bare alef}, \textit{alef maqsoura} to \textit{ya}, and \textit{ta marbouta} to \textit{ha}.  Table \ref{tab:baselineCrossValResults} reports on the baseline results.  The results show that though SVM yielded slightly higher precision than fastText, but fastText had much higher recall leading to significantly higher F1 measure.  \\
\textbf{Training Set Expansion:} We have two main goal for this work, namely: to expand the offensive language training to have broad coverage; and to identify target specific offensive language.  Doing so would hopefully improve overall classification effectiveness.  \\
\begingroup
\renewcommand{\arraystretch}{1.2} 
\begin{table*}[!t]
    \small
    \centering
    \begin{tabular}{l|l|r||l|l|r}
Accounts	&	Description	&	Replies & Accounts	&	Description	&	Replies	\\ \hline
AA\_Arabic	&	Anadolu Agency -- news agency	&	1,733	& 
AbdullahElshrif	&	Egyptian youtuber/dissident	&	10,036	\\
AJArabic	&	Aljazeera Arabic -- TV channel	&	10,316	&
AlArabiya	&	TV channel	&	5,633	\\
AlSisiOfficial	&	Egyptian politician	&	11,031	&
AmrWaked	&	Egyptian actor/dissident	&	11,136	\\
BakryMP	&	Egyptian politician/TV presenter	&	13,061	&
Dhahi\_Khalfan	&	UAE public official	&	13,592	\\
DrAlaaSadek	&	Egyptian TV commentator	&	2,235	&
EdyCohen	&	Israeli journalist	&	10,004	\\
ElWatanNews	&	Egyptian newspaper	&	3,047	&
GamalEid	&	Egyptian political dissident	&	2,365	\\
Gebran\_Bassil	&	Lebanese politician	&	1,583	&
IsraelArabic	&	Israeli government media	&	2,404	\\
JaberAlharmi	&	Qatari journalist	&	4,855	&
JamalRayyan	&	Aljazeera presenter	&	5,881	\\
KalNaga	&	Egyptian actor/dissident	&	3,773	&
KasimF	&	Aljazeera presenter	&	7,817	\\
MAlHachimi	&	Tunisian politician	&	3,336	&
NBenotman	&	Libyan public personality	&	2,250	\\
OfficialAmro1	&	Egyptian Twitter account	&	6,253	&
RTArabic	&	Russia Today -- news	&	11,812	\\
TawakkolKarman	&	Yemeni Nobel prize lauret	&	5,721	&
TurkiShalhoub	&	Saudi dissident	&	8,775	\\
Youm7	&	Egyptian newspaper	&	6,877	& & & \\
    \end{tabular}
    \caption{25 accounts with the most number of offensive replies with the number of their replies.}
    \label{tab:accountsWithMostOffReplies}
\squeezeupless
\end{table*}
\endgroup

\textbf{Test Set:} To assess the efficacy of the expanded training data and its effect on specific targets, we identified 25 target entities that were replied to the most in the offensive tweets training set.  For example, in the tweet: @BakryMP
\<بس يا متخلف يا مطبلاتي يا كلب > 
(``bs yA mtxlf yA mTblAty yA klb -- enough you idiot, you propagandist, you dog''), the target is Mostafa Bakry (``@BakryMP''), an Egyptian media personality and politician. For each account, we obtained all replies to them from a set of 660k Arabic tweets that we collected between April 15 -- May 6, 2019.  Table \ref{tab:accountsWithMostOffReplies} lists the 25 target accounts.  For each account, we randomly selected 100 random replies, and we manually annotated them for offensiveness. Offensiveness was predicated on being directed towards the target entity. Thus, an offensive tweet that is not targeted at the specific entity was not considered offensive. In all, we manually annotated 2,500 tweets from 25 accounts, which we use for testing.  We plan to release all the annotated test tweets.\\

\begin{table*}[!htb]
\captionsetup{justification=centering}
 \centering
    \begin{subtable}{.52\linewidth}
      \centering
    \footnotesize
\begin{tabular}{c|c|c|c|c|c|c}
    & \multicolumn{3}{c|}{fastText} & \multicolumn{3}{c}{SVM} \\
	&	P	&	R	&	F1	&	P	&	R	&	F1	\\ \hline
baseline	&	\textbf{79.0}	&	56.8	&	65.6	&	\underline{\textbf{92.5}}	&	19.1	&	31.1	\\ \hline
50\%	&	69.6	&	80.5	&	73.2	&	90.9	&	23.7	&	36.7	\\
top 10	&	73.6	&	70.2	&	70.9	&	86.6	&	30.3	&	43.9	\\
top 20	&	71.3	&	75.8	&	72.4	&	84.5	&	35.2	&	48.6	\\
top 50	&	68.7	&	\underline{\textbf{83.4}}	&	\underline{\textbf{74.1}}	&	80.4	&	\textbf{44.1}	&	\textbf{55.6}	\\
    \end{tabular}
        \caption{}
        \label{tab:perTopicExpansionResults}
        \squeezeupless
    \end{subtable}%
    \begin{subtable}{.52\linewidth}
      \centering
   \footnotesize
            \begin{tabular}{c|c|c|c|c|c|c}
    & \multicolumn{3}{c|}{fastText} & \multicolumn{3}{c}{SVM} \\
	&	P	&	R	&	F1	&	P	&	R	&	F1	\\ \hline
baseline	&	\textbf{83.9}	&	65.4	&	\textbf{73.5}	&	89.7	&	36.0	&	51.4	\\\hline
50\%	&	65.8	&	\underline{\textbf{70.9}}	&	68.2	& \underline{\textbf{96.4}}	&	64.0	&	76.9	\\
top 10	&	72.8	&	69.8	&	71.2	&	94.7	&	64.5	&	76.7	\\
top 20	&	68.4	&	70.2	&	69.2	&	93.4	&	64.1	&	76.0	\\
top 50	&	65.9	&	70.8	&	68.2	&	92.5	& \textbf{64.7}	&	\underline{\textbf{76.1}}	\\
    \end{tabular}
        \caption{} \label{tab:crossValidResultsExpansion}
        \squeezeupless
    \end{subtable} 
    \caption{Results of (a) per target expansion and (b) overall training set expansion. Best values for P, R and F1 are underlined. Baseline results are obtained when we train  on original data without expansion.}
    \label{tab:baselineCrossValResults}
\squeezeup
\end{table*}

\begin{table}[!]
    \footnotesize
    \centering
    \begin{tabular}{l|c|c|c|c|}
	&	50\%	&	top 10	&	top 20	&	top 50	\\ \hline

fastText Avg	&	799	&	189	&	299	&	539	\\ \hline
SVM Avg	&	54	&	132	&	208	&	366
	\\
    \end{tabular}
    \caption{Average number of expansion tweets per target}
    \label{tab:countOfExpansionTweets}
\squeezeup
\end{table}
\squeezeupless

\subsection{Expansion Experiments} 
For expansion, we borrowed a technique from information retrieval called \textit{blind relevance feedback} \cite{salton1990improving}, which involves: 1) using an initial query to search a document collection; 2) assuming the top returned documents are relevant; and 3) using these documents to reformulate the query, hopefully leading to better results. Analogously, for users who frequently reply to a target and a classifier deems that some portion of their replies are offensive, we assumed that the rest of their replies are also offensive.  The intuition behind this is that users typically have relative stable stances over time (antagonistic in this case) \cite{borge2015content}.  Further, if the classifier identified some of the replies of a user as offensive, it is likely that the user's other replies were also offensive and the classifier missed them. Thus, given the 25 accounts in Table \ref{tab:accountsWithMostOffReplies}, we automatically tagged all tweets that reply to them using our aforementioned baseline classifiers.  Then, we identified the ``most offensive'' users who either: 1) had at least 50\% of their replies automatically tagged as offensive (\textit{50\%} users); or 2) were in the top \textit{n} users with most number of offensive replies, where we experimented with values of \textit{n} = 10, 20, and 50 (\textit{top n} users).  We used the tweets of these users to expand our training set.
\section{Results and Discussions}
We conducted two types of experiments.  In the first, for each target, we added the tweets of the most offensive users for that target to the training set, and we retrained the classifiers.  Then, we tested on the 100 manually labeled replies for each target.  Table \ref{tab:countOfExpansionTweets} lists the average number of expansion tweets for our targets.  Not surprisingly, fastText tended to yield more expansion tweets, as the baseline fastText classifier generally had higher recall compared to SVM. Table \ref{tab:perTopicExpansionResults} 
reports on the average precision, recall, and F1-measure across all topics when using the fastText and SVM classifiers with and without expansion. 
Though expansion generally lowered precision, it greatly improved recall leading to significantly higher F1-measure.  For per target results, F1 scores for any of the expansion setups was rarely lower than the baseline (no expansion).  Only one target, namely @AlSisiOfficial, was significantly adversely affected by expansion. We found that many of the replies were in fact Sisi supporters who were attacking his enemies. Using the \textit{top 50} users for expansion led to the most improvement in F1-measure for both classifiers. In the case of fastText, F1 increased from 65.6 to 74.1 (12.9\% relative improvement), and for SVM, it increased from 31.1 to 55.6 (78.8\% relative improvement).

In the second set of experiments, we used cross-validation on the training set, where we enriched the training folds with all the expansion tweets of all of the most offensive users across all target entities.  We identified the most offensive users using the training folds only.  
Then, we tested on the test folds (SemEval 2020).  Table \ref{tab:crossValidResultsExpansion} reports the results with and without expansion.  The results show that the effect of expansion on fastText and SVM varied significantly.  For fastText, precision dropped significantly with expansion leading to modest improvement in recall, yielding to lower F1-measure.  However, SVM-based expansion improved both precision and recall, leading to an overall improvement of 48.1\% in F1-measure.  We suspect that fastText performed worse than SVM due to the large class imbalance that resulted from adding the tweets of the offensive users.  Prior to expansion with fastText, the ratio of non-offensive to offensive tweets was 4 to 1, and became 1 to 2 for the 50\% users after expanding.  Further, SVM initially had higher precision compared to fastText, leading to more prudent expansion.



\section{Conclusion}
In this paper, we show the effectiveness of automatically expanding offensive language training set with target specific tweets.  The approach yields improved results for identifying target-specific offensive tweets in specific and for identifying offensive tweets in general. Our approach is based on that intuitions that replies are often used for attacks and Twitter users that write offensive language replies directed towards an entity are likely to have authored other offensive replies.  We applied our work on Arabic offensive language tweets and used SVM-based and deep-learning based classifiers in our work. SVM-based classification generally favored precision, while the latter favored recall.  For future work, we plan to apply this technique to other languages, and to use transformer-based representations in classification.

\bibliography{anthology,custom}
\bibliographystyle{acl_natbib}




\end{document}